\documentclass[copyright,creativecommons]{eptcs}
\providecommand{\event}{}
\providecommand{\volume}{??}
\providecommand{\anno}{20??}
\providecommand{\firstpage}{1}
\usepackage{breakurl}
\usepackage[pdftex]{graphicx}

\title{How creative should creators be to optimize the evolution of ideas? A computational model}
\author{Stefan Leijnen
\institute{University of British Columbia\\
Okanagan campus, 3333 University Way \\
Kelowna BC, V1V 1V7, CANADA}
\email{stefanleijnen@gmail.com}
\and
Liane Gabora
\institute{University of British Columbia\\
Okanagan campus, 3333 University Way \\
Kelowna BC, V1V 1V7, CANADA}
\institute{and PACE Center, Tufts University}
\email{liane.gabora@ubc.ca}
}
\def\titlerunning{How creative should creators be to optimize the evolution of ideas?}
\def\authorrunning{L. Gabora \& S. Leijnen}
\begin{document}
\maketitle

\begin{abstract}
There are both benefits and drawbacks to creativity. In a social group it is not necessary for all members to be creative to benefit from creativity; some merely imitate or enjoy the fruits of others' creative efforts. What proportion should be creative? This paper contains a very preliminary
 investigation of this question carried out using a computer model of cultural evolution referred to as EVOC (for EVOlution of Culture). EVOC is composed of neural network based agents that evolve fitter ideas for actions by (1) inventing new ideas through modification of existing ones, and (2) imitating neighbors' ideas. The ideal proportion with respect to fitness of ideas occurs when thirty to forty percent of the individuals is creative. When creators are inventing 50\% of iterations or less, mean fitness of actions in the
society is a positive function of the ratio of creators to imitators; otherwise
mean fitness of actions starts to drop when the ratio of creators to imitators
 exceeds approximately 30\%. For all levels of creativity, the diversity of ideas in a population is positively correlated with the ratio of creative agents.
\end{abstract}

\section{Introduction}

Computer science is drawing ever more extensively upon the natural world for inspiration in the design of search algorithms, optimization tools, problem solving techniques, and even computer-based artistic expression. Probably the most effective problem solver Mother Nature has come up with is the human mind itself. Its effectiveness derives largely from the fact that it is endlessly creative, able to break out of ruts and come up with ideas and solutions that are new, useful, and appealing. Not only are we individually creative, but we build on each other's creations such that over the centuries our ideas and inventions can be said to have evolved. In order for computer scientists to put to use the process by which creative ideas evolve through cultural exchange we must first get a deeper computational understanding of it. This paper presents investigations of one aspect of the process: the interaction between how creative individuals are, and how numerous they are in a society. \\
That human creativity is immensely beneficial is fairly obvious. Our considerable capacity for self-expression, for finding practical solutions to problems of survival, and coming up with aesthetically pleasing objects that delight the senses, all stem from the creative power of the human mind. However, there are also considerable drawbacks to creativity. A creative solution to one problem often generates other problems or unexpected negative side effects that may only become apparent after much has been invested in the creative solution. Moreover, creative individuals are more emotionally unstable and prone to affective disorders such as depression and bipolar disorder, and have a higher incidence of schizophrenic tendencies, than other segments of the population (Andreason, 1987; Flaherty, 2005; Jamieson, 1989, 1993; Styron, 1990). They are also more prone to abuse drugs and alcohol (Goodwin, 1988, 1992; Ludwig, 1995; Norlander \& Gustafson, 1996, 1997, 1998; Rothenberg, 1990) as well as suicide (Goodwin \& Jamieson, 1990). Also, creative people often feel disconnected from others because they defy the crowd (Sternberg \& Lubart, 1995; Sulloway, 1996). However, in a group of interacting individuals only a fraction of them need be creative for the benefits of creativity to be felt throughout the group. The rest can reap the benefits of the creator's ideas without having to withstand the dark aspects of creativity by simply copying, using, or admiring them. After all, few of us know how to build a computer, or write a symphony, or a novel, but they are nonetheless ours to use and enjoy when we please. One can thus ask: in order for a culture to evolve optimally, what proportion of individuals should be 'creative types' and how creative should they be? \\
This paper investigates this using an agent-based modeling approach. The agents are too simple to develop affective disorders or abuse alcohol; the drawback to their creativity is that complex solutions to multi-part problems (in biological terms, epistatic relations) break down when too much variation is introduced too quickly. Ideas that are \emph{too} creative are not implemented as actions in the world; thus if an agent is \emph{overly} creative, its creative potential is wasted. Moreover, since each iteration each agent either invents or imitates, by deciding (and failing) to invent, it forgoes a chance to imitate.

\section{The Modeling Approach}

EVOC consists of neural network based agents that invent ideas for actions, and imitate neighbors' actions Gabora, 2008). EVOC is an elaboration of Meme and Variations, or MAV (Gabora, 1995), the earliest computer program to model culture as an evolutionary process in its own right. MAV was inspired by the genetic algorithm (GA), a search technique that finds solutions to complex problems by generating a 'population' of candidate solutions through processes akin to mutation and recombination, selecting the best, and repeating until a satisfactory solution is found. Although MAV has inspired the incorporation of cultural phenomena (such as imitation, knowledge-based operators, and mental simulation) into evolutionary search algorithms (e.g. Krasnogor \& Gustafson, 2004), the goal behind MAV was not to solve search problems, but to gain insight into how ideas evolve. It used neural network based agents that could (1) invent new ideas by modifying previously learned ones, (2) evaluate ideas, (3) implement ideas as actions, and (4) imitate ideas implemented by neighbors. Agents evolved in a cultural sense, by generating and sharing ideas for actions, but not in a biological sense; they neither died nor had offspring. The approach can thus be contrasted with computer models of the interaction between biological evolution and individual learning (Best, 1999, 2006; Higgs, 2000; Hinton \& Nowlan, 1987; Hutchins \& Hazelhurst, 1991). \\
MAV successfully modeled how 'descent with modification' can occur in a cultural context, but it had limitations arising from the outdated methods used to program it. Moreover, although new ideas in MAV were generated making use of acquired knowledge and pattern detection, the name 'Meme and Variations' implied acceptance of the notion that cultural novelty is generated randomly, and that culture evolves through a Darwinian process operating on discrete units of culture, or 'memes'. Problems with memetics and other Darwinian approaches to culture have become increasingly apparent (Boone \& Smith, 1998; Fracchia \& Lewontin, 1999; Gabora, 2004, 2006, 2008; Jeffreys, 2000). One problem is that natural selection prohibits the passing on of acquired traits (thus you don't inherit your mother's tattoo).\footnote{That isn't to say that inheritance of acquired traits never occurs in biological evolution; it does. However to the extent that this is the case natural selection cannot provide an accurate model of biological evolution. Because inheritance of acquired traits is the exception in biology not the rule, natural selection still provides a roughly accurate model of biological evolution.} In culture, however, 'acquired' change-that is, modification to ideas between the time they are learned and the time they are expressed-is unavoidable. Darwinian approaches must assume that elements of culture are expressed in the same form as that in which they are acquired. Natural selection also assumes that lineages do not intermix. However, because ideas cohabit a distributed memory with a multitude of other ideas, they are constantly combining to give new ideas, and their meanings, associations, and implications are constantly revised. It has been proposed that what evolves through culture is not discrete memes or artifacts, but the internal models of the world that give rise to them (Gabora, 2004), and they evolve not through a Darwinian process of competitive exclusion but a Lamarckian process involving exchange of innovation protocols (Gabora, 2006, 2008). EVOC incorporates this in part by allowing agents to have multiple interacting needs, thereby fostering complex actions that fulfill multiple needs. Elsewhere (Gabora, 2008a,b) results of experiments using different needs and/or multiple needs are described, as well as how cultural evolution is affected by affordances of the agents' world, such as world shape and size, population density, and barriers that impede information flow, and potentially erode with time. This paper investigates how different proportions of creative to uncreative agents affects the fitness and diversity of ideas.

\section{Architecture}

EVOC consists of an artificial society of agents in a two-dimensional grid-cell world. This section describes the key components of the agents and the world they inhabit.

\subsection{The Agent}

Agents consist of (1) a neural network, which encodes ideas for actions and detects trends in what constitutes a fit action, and (2) a body, which implements actions. In MAV there was only one need-to attract a mate. Thus actions were limited to gestures that attract mates. In EVOC agents can also engage in tool-making actions.

\subsubsection{The Neural Network}

\begin{figure}[htp]
\centering
\includegraphics[width=80mm]{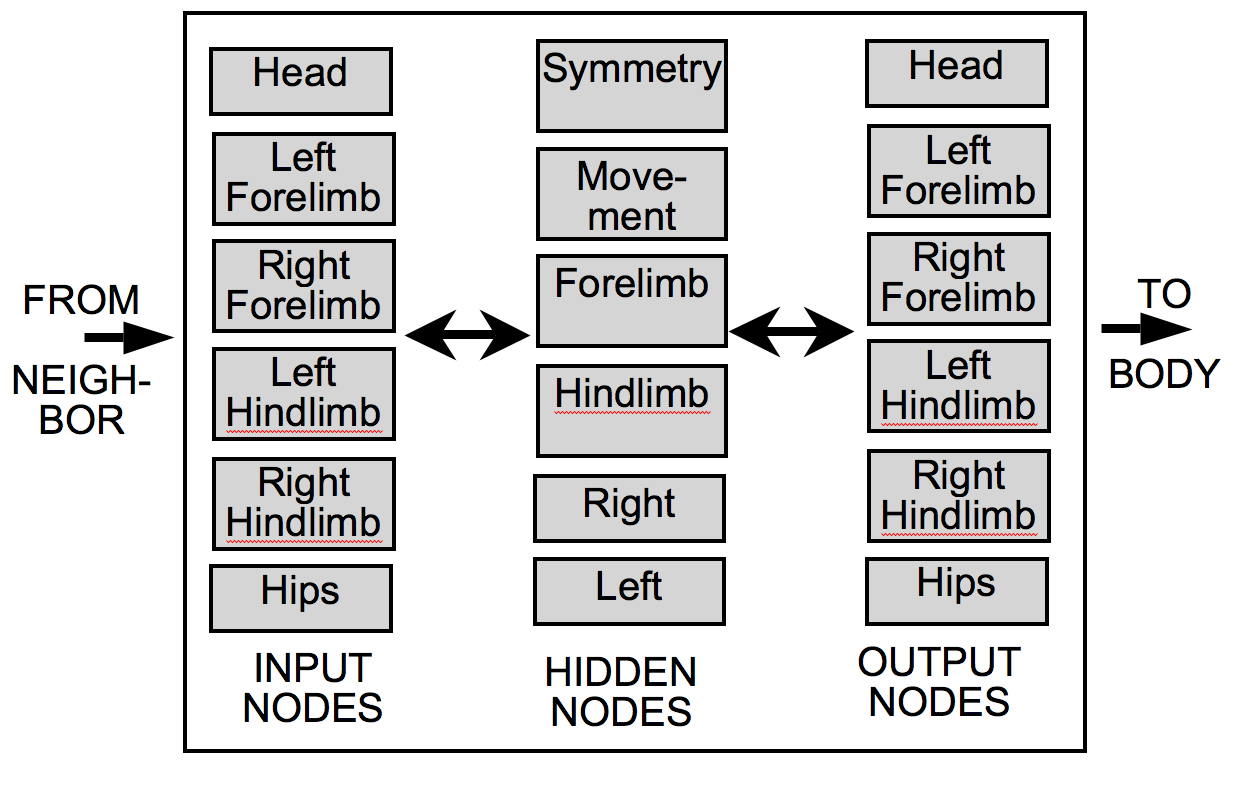}
\caption{The neural network. See text for details.}
\label{fig:neuralnetwork}
\end{figure}

The core of an agent is a neural network, as shown in Figure~\ref{fig:neuralnetwork}. It is composed of six input nodes that represent concepts of body parts (LEFT ARM, RIGHT ARM, LEFT LEG, RIGHT LEG, HEAD, and HIPS), six matching output nodes, and six hidden nodes that represent more abstract concepts (LEFT, RIGHT, ARM, LEG, SYMMETRY and MOVEMENT). Input nodes and output nodes are connected to 'hidden' nodes of which they are instances (e.g. RIGHT ARM is connected to RIGHT.) Activation of any input node increases activation of the MOVEMENT hidden node. Opposite-direction activation of pairs of limb nodes (e.g. leftward motion of one arm and rightward motion of the other) activates the SYMMETRY node. The neural network learns ideas for actions. An idea is a pattern of activation across the output nodes consisting of six elements that instruct the placement of the six body parts. Training of the neural network is as per (Gabora, 1995). In brief, the neural network starts with small random weights, and patterns that represent ideas for actions are presented to the network. Each time a pattern is presented, the network's actual output is compared to the desired output. An error term is computed, which is used to modify the pattern of connectivity in the network such that its responses become more correct. Since the neural network is an autoassociator, training continues until the output is identical to the input. At this point training stops and the run begins. The value of using a neural network is simply that trends about what makes for a fit action can be detected using the symmetry and movement nodes (see below). The neural network can also be turned off to compare results to those obtained using instead of a neural network a simple data structure that cannot detect trends, and thus invents ideas at random.

\subsubsection{Knowledge-based Operators}

Brains detect regularity and build schemas with which they adapt the mental equivalents of mutation and recombination to tailor actions to the situation at hand. Thus they generate novelty strategically, on the basis of past experience. Knowledge-based operators are a crude attempt to incorporate this into the model. Since a new idea for an action is not learned unless it is fitter than the currently implemented action, newly learned actions provide valuable information about what constitutes an effective idea. This information is used by knowledge-based operators to probabilistically bias invention such that new ideas are generated strategically as opposed to randomly. Thus the idea is to translate knowledge acquired during evaluation of an action into educated guesses about what makes for a fit action. Two rules of thumb are used. The first rule is: if movement is generally beneficial, the probability increases that new actions involve movement of more body parts. Each body part starts out at a stationary rest position, and with an equal probability of changing to movement in one direction or the other. If the fitter action codes for more movement, increase the probability of movement of each body part. Do the opposite if the fitter action codes for less movement.\\
This rule of thumb is based on the assumption that movement in general (regardless of which particular body part is moving) can be beneficial or detrimental. This seems like a useful generalization since movement of any body part uses energy and increases the likelihood of being detected. It is implemented as follows:

\begin{verbatim}

am1 = movement node activation for current action
am2 = movement node activation for new action
p(im)i = probability of increased movement at body part i
p(dm)i = probability of decreased movement at body part i

IF (am2 >  am1)
THEN p(im)i = MAX(1.0, p(im)i + 0.1)
ELSE IF (am2 <  am1)
THEN p(im)i = MIN(0.0, p(im)i - 0.1)
p(dm)i = 1 - p(im)i

\end{verbatim}

\noindent The second rule of thumb is: if fit actions tend to be symmetrical (e.g. left arm moves to the right and right arm moves to the left), the probability increases that new actions are symmetrical. This generalization is biologically sensible, since many useful actions (e.g. walking) entail movement of limbs in opposite directions, while others (e.g. pushing) entail movement of limbs in the same direction. This rule is implemented in a manner analogous to that of the first rule. In summary, each action is associated with a measure of its effectiveness, and generalizations about what seems to work and what does not are translated into guidelines that specify the behavior of the algorithm.

\subsubsection{The Body}

If the fitness of an action is evaluated to be higher than that of any action learned thus far, it is copied from the output nodes of the neural network that represent concepts of body parts to a six digit array that contains representions of actual body parts, referred to as the body. Since it is useful to know how many agents are doing essentially the same thing, when node activations are translated into limb movement they are thresholded such that there are only three possibilities for each limb: stationary, left, or right. Six limbs with three possible positions each gives a total of 729 possible actions. Only the action that is currently implemented by an agent's body can be observed and imitated by other agents.

\subsection{The Fitness Functions}

Agents evaluate the effectiveness of their actions according to how well they satisfy needs using a pre-defined equation referred to as a fitness function. Agents have two possible needs. The fitness of an action with respect to the need to attract mates is referred to as F1, and it is calculated as in (Gabora, 1995). F1 rewards actions that make use of trends detected by the symmetry and movement hidden nodes and used by knowledge-based operators to bias the generation of new ideas. F1 generates actions that are relatively realistic mating displays, and exhibits a cultural analog of epistasis. In biological epistasis, the fitness conferred by the allele at one gene depends on which allele is present at another gene. In this cognitive context, epistasis is present when the fitness contributed by movement of one limb depends on what other limbs are doing. In these simulations F1 is used exclusively.

\subsection{Incorporation of Cultural Phenomena}

In addition to knowledge-based operators, discussed previously, agents incorporate the following phenomena characteristic of cultural evolution as parameters that can be turned off or on (in some cases to varying degrees):
\begin{itemize}
\item Imitation Ideas for how to perform actions spread when agents copy neighbors' actions. This enables them to share effective, or 'fit', actions.
\item Invention. This code enables agents to generate new actions by modifying their initial action or a previously invented or imitated action using knowledge-based operators (discussed previously).
\item Mental simulation. Before implementing an idea as an action, agents can use the fitness function to assess how fit the action would be if it were implemented.
\end{itemize}

\subsection{The World}

MAV allowed only worlds that were toroidal, or 'wrap-around'. Moreover, the world was always maximally densely populated, with one stationary agent per cell. In EVOC the world can be either toroidal or square, and as sparsely or densely populated as desired, with stationary agents placed in any configuration. EVOC also allows the creation of complete or semi-permeable borders, which may be permanent or eroding. (This limits the probability of agents imitating others from different enclaves.)

\subsection{A Typical Run}

Each iteration, every agent has the opportunity to (1) acquire an idea for a new action, either by imitation, copying a neighbor, or by invention, creating one anew, (2) update the knowledge-based operators, and (3) implement a new action. To invent a new idea, the current action is copied to the input layer of the neural network, and this previous action is used as a basis from which to generate a new one. For each node the agent makes a probabilistic decision as to whether change will take place. If it does, the direction of change is stochastically biased by the knowledge-based operators using the activations of the SYMMETRY and MOVEMENT nodes. Mental simulation is used to determine whether the new idea has a higher fitness than the current action. If so, the agent learns and implements the action specified by the new idea. To acquire an idea through imitation, an agent randomly chooses one of its neighbors, and evaluates the fitness of the action the neighbor is implementing using mental simulation. If its own action is fitter than that of the neighbor, it chooses another neighbor, until it has either observed all of its immediate neighbors, or found one with a fitter action. If no fitter action is found, the agent does nothing. Otherwise, the neighbor's action is copied to the input layer, learned, and implemented. Fitness of actions starts out low because initially all agents are immobile. Soon some agent invents an action that has a higher fitness than doing nothing, and this action gets imitated, so fitness increases. Fitness increases further as other ideas get invented, assessed, implemented as actions, and spread through imitation. The diversity of actions initially increases due to the proliferation of new ideas, and then decreases as agents hone in on the fittest actions.

\subsection{Implementation}

EVOC is written in Java, an object oriented programming environment, using the Joone open source neural network library. The graphical user interface makes use of the open-source charting project, JFreeChart, enabling variables to be user defined at run time, and results to become visible as the computer program runs.

\section{Summary of Previous Results}
\begin{figure}[htp]
\centering
\includegraphics[width=160mm]{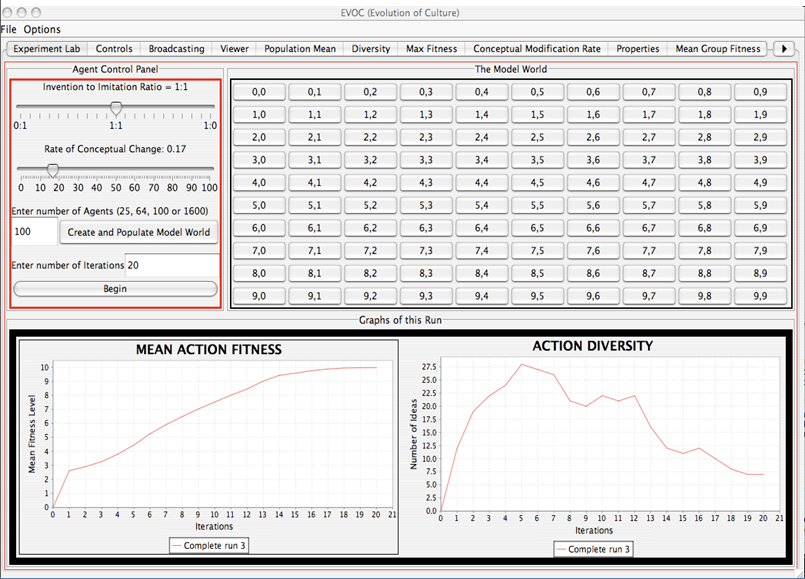}
\caption{Output panel of GUI using F$_{\textrm{1}}$. See text for details.}
\label{fig:evoc}
\end{figure}

EVOC closely replicates the results of experiments conducted with MAV (Gabora, 1995). The graph on the bottom left of Figure~\ref{fig:evoc} shows the increase in fitness of actions. The graph on the bottom right of Figure~\ref{fig:evoc} shows the increase and then decrease in the diversity of actions. Other results include:

\begin{itemize}
\item Fitness increases most quickly with an invention to imitation ratio of about 2:1.
\item For the agent with the fittest actions, however, the less it imitates, the better it does.
\item Increasing the invention-to-imitation ratio increases the diversity of actions. If increased much beyond 2:1, it takes more than twice as many iterations for all agents to settle on optimal actions.
\item As in biology, epistatically linked elements take longer to optimize. (As explained earlier, in the present context epistasis refers to the situation where the effect on fitness of what one limb is doing depends on what another is doing.)
\item The program exhibits drift-the biological term for change in the relative frequencies of alleles (forms of a gene) as a statistical byproduct of randomly sampling from a finite population (Wright, 1969). With respect to culture, it pertains to possible forms of a component of an idea (e.g. if the idea is to implement the gesture 'wave', this can be executed with the left or right hand).
\item Diversity of actions is positively correlated with number of needs that agents attempt to satisfy.
\item Diversity of actions is positively correlated with population size and density, and with barriers between populations.
\item Square (as opposed to toroidal) worlds foster higher diversity, as idea propagation is impeded by corners and edges.
\item Slowly eroding borders increase fitness without sacrificing diversity by fostering specialization followed by sharing of fit actions.
\item Introducing a leader that broadcasts its actions throughout the population increases the fitness of actions but reduces diversity of actions. Increasing the number of leaders reduces this effect.
\end{itemize}

\section{Experiments}

In previous experiments (Gabora, 1995, 2008b, 2008c), all agents have an equal probability of inventing or imitating. The choice of action is determined by a number of different factors and the role an agent may take can vary each time step. In these simulation experiments we make a distinction between two types of agents, which differ in the extent to which they are creative, i.e. able to invent new actions by modifying previous ones. Whereas one kind of agent, referred to as imitators, always obtains new actions by imitating neighbors, the other type of agent, referred to as inventors or creators, will generally obtain new actions by inventing them. 

\begin{figure}[htp]
\centering
\includegraphics[width=160mm]{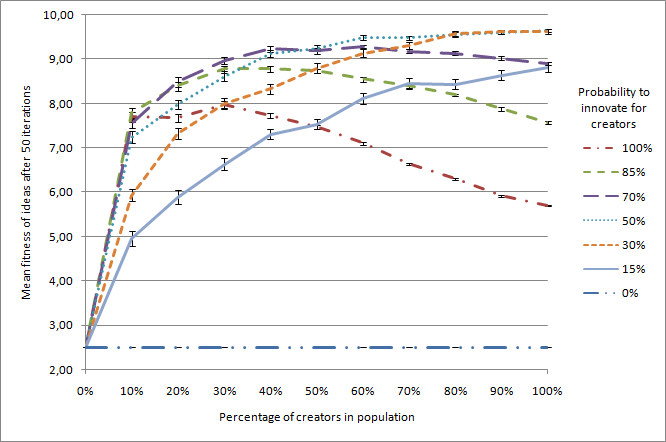}
\caption{Effect of increasing percentage of creators in the population on mean fitness of ideas at 15 iterations, for different degrees to which creators are creative.}
\label{fig:fitness}
\end{figure}

\noindent In these simulations there are two negative consequences of creativity. The first is that an iteration spent inventing is an iteration not spent imitating. The second is that creative change can break up co-adapted partial solutions. Actions have a cultural version of what in biology is referred to as epistasis, wherein what is optimal with respect to one part depends on what is done with respect to another part. Once both parts of the problem have been solved in a mutually beneficial way, too much creativity can cause these co-adapted solutions to break down.\\
In the experiments reported here, the world used for the simulation is toroidal and contains 100 cells with one agent in each cell, and the world is not segmented, i.e. there are no barriers as in other EVOC experiments reported elsewhere. Broadcasting is not used, and there is a 1/6\% probability\footnote{With six body parts, on average each newly invented action differs from the one it was based on with respect to one body part (cf. Gabora, 2008c)} of change to any body part during innovation. All experiment results displayed in graphs are averaged over 100 runs; on each run, the creative agents are randomly dispersed.

\subsection{Proportion of Creators to Imitators}

\begin{figure}[htp]
\centering
\includegraphics[width=160mm]{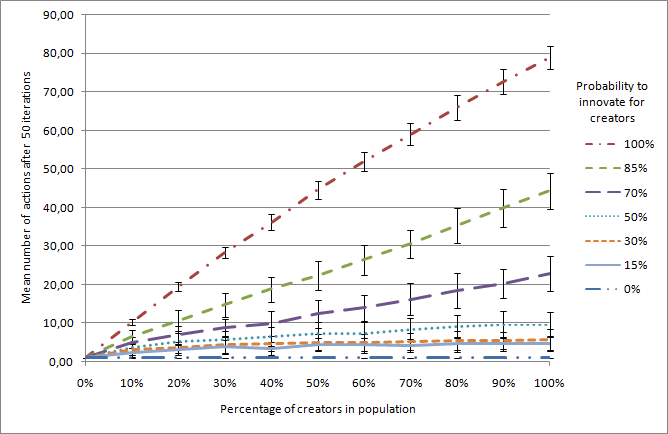}
\caption{Effect of increasing percentage of creators in the population on diversity of ideas at 15 iterations, for different degrees to which creators are creative.}
\label{fig:diversity}
\end{figure}

The first experiment investigates the effect of varying the ratio of creative agents to imitators on mean fitness of actions across the society. In these simulations, all actions are generated by the creative agents; that is, imitators do not invent at all. Imitators simply copy the successful innovations of the creative agents, and thereby serve as a 'memory' for preserving the fittest configurations. In different runs we not only observed the effect of changing how rare creative agents are, but how creative they are. That is, in some runs they not only invent but also imitate, and the ratio of invention to imitation was varied across different runs. \\
Figure~\ref{fig:fitness} shows that increasing the percentage of creators in a population has a positive effect on the mean action fitness, when the creators invent new ideas half or less than half of the time. If the creative agents invent more often, the optimal ratio decreases to approximately 30-40\% of the population. When the whole of society consists entirely of imitators, no new actions are created and the fitness remains the same. Figure~\ref{fig:diversity} illustrates that the number of different actions in the artificial society is positively correlated with the percentage of creators, for all levels of creativity. As the proportion of creators increases, a larger fraction of the search space is discovered.

\section{Discussion}

It is known that creativity has drawbacks as well as benefits. The goal of the work reported here was to investigate when (if ever) is creativity too much of a good thing. In the experiments reported here there is no possibility of newly invented ideas having unexpected negative side-effects or consequences. Nor is creativity associated with affective disorders such as depression. There are only two negative consequences of creativity in the simulations. The first is that an iteration spent inventing is an iteration not spent imitating. The second is that creative change can break up co-adapted partial solutions.\\
In previous work it was found that the ideal ratio of imitating to inventing was approximately 2:1 (Gabora, 1995). That is, the fitness of actions evolved most quickly when in 2/3 of iterations agents invented and in 1/3 of iterations agents imitated. In these previous experiments all agents were equally capable of both inventing and imitating. In the current experiments investigate a related but different question: what proportion of individuals should be 'creative types'? The rationale is that it is known that in a society of interacting individuals capable of imitation, some members can capitalize on the benefits of creativity without incurring the drawbacks by merely imitating their creative peers. So if only some fraction of the population are creators, and the rest imitators, what is the ideal ratio of creators to imitators?

We found that when creators innovate more than 50\% of the time, and imitators imitate 100\% of the time, the ideal proportion is approximately 30-40\% creators and 60-70\% imitators. As the creativity of the creators decreases, their ideal number increases up to a hundred percent. In general, the more creative the creators are, the less numerous they should be.

\section*{Acknowledgments}
This research is supported by grants to Liane Gabora from the Social Sciences and Humanities Research Council of Canada and the Flemish Government of Belgium's Concerted Research Program.

\bibliographystyle{eptcs} 

\end{document}